\pdfoutput=1
\documentclass[11pt]{article}

\usepackage[final]{acl}

\usepackage{times}
\usepackage{latexsym}
\usepackage{booktabs}
\usepackage{graphicx}

\usepackage[T1]{fontenc}
\usepackage[utf8]{inputenc}

\usepackage{microtype}

\usepackage{inconsolata}

\usepackage{graphicx}

\title{Regulation of Language Models With Interpretability Will Likely Result In A Performance Trade-Off}

\author{Eoin M. Kenny \\
  CSAIL \\
  Massachusetts Institute of Technology \\
  Cambridge, MA, U.S.A. \\
  \texttt{ekenny@mit.edu} \\\And
  Julie A. Shah \\
  AeroAstro \\
  Massachusetts Institute of Technology \\
  Cambridge, MA, U.S.A. \\
  \texttt{arnoldj@mit.edu} \\}

\begin{document}
\maketitle
\begin{abstract}
Regulation is increasingly cited as the most important and pressing concern in machine learning.
However, it is currently unknown how to implement this, and perhaps more importantly, how it would effect model performance alongside human collaboration if actually realized.
In this paper, we attempt to answer these questions by building a regulatable large-language model (LLM), and then quantifying how the additional constraints involved affect (1) model performance, alongside (2) human collaboration.
Our empirical results reveal that it is possible to force an LLM to use human-defined features in a transparent way, but a ``regulation performance trade-off'' previously not considered reveals itself in the form of a 7.34\% classification performance drop.
Surprisingly however, we show that despite this, such systems actually improve human task performance speed and \textit{appropriate} confidence in a realistic deployment setting compared to no AI assistance, thus paving a way for fair, regulatable AI, which benefits users.\footnote{Code is available at \url{https://github.com/EoinKenny/Regulatable_LLMs}}
\end{abstract}

\section{Introduction}
Ineffective regulation of AI and the neglection of safety is often cited as the biggest existential threat to humanity~\cite{bengio2024managing}.
Ex-board members of OpenAI have recently been quoted as saying governments must begin building effective regulatory frameworks \textit{now}, as AI firms cannot self-govern and reliably withstand the pressure of profit incentives~\cite{toner_mccauley_2024}.
The biggest factor pushing this regulatory interest is large-language models (LLMs)~\cite{vaswani2017attention}, which have already had far reaching consequences in society, ranging from medicine to self-driving cars~\cite{chen2023driving}.
The core issue is that these systems cannot escape the same limitation that underlines most neural network architectures, in that they are black boxes with no obvious interpretable decision-making process, making it completely impossible to use or audit them for any sensitive application~\cite{keaneif}.
Governments at large are aware of this and the European AI Act is a sign of things to come in how they will continue to heavily regulate AI both in Europe and North America~\cite{smuha2021eu}.
However, it is presently unclear how LLMs might be regulated in practice.

\begin{figure*}[t!]
  \includegraphics[width=\textwidth]{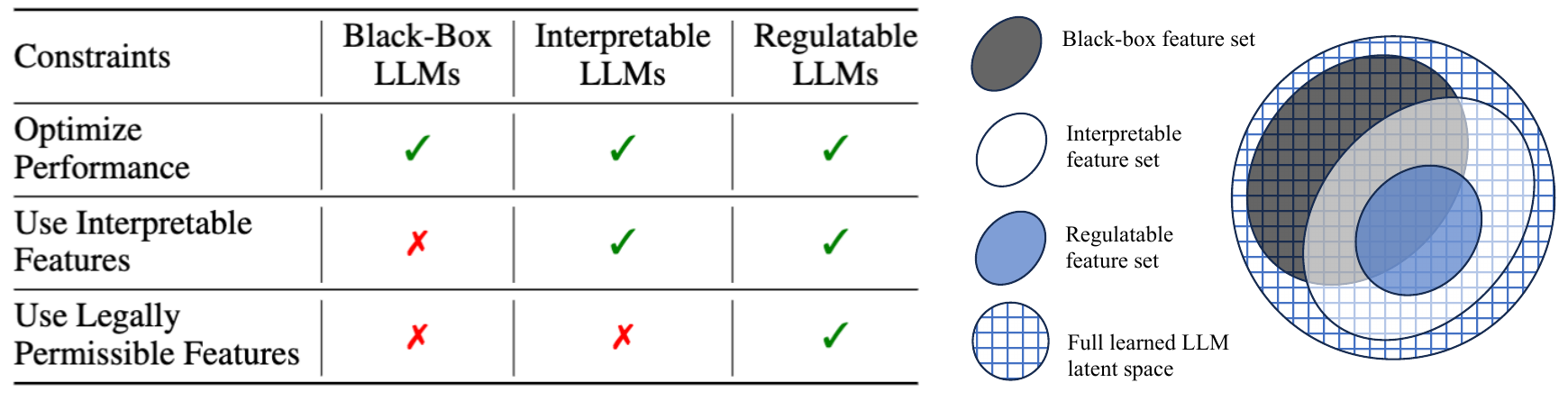}
  \caption{
  The Regulation Performance Trade-Off: 
  A black-box LLM will learn to use the optimal feature set which minimizes its loss on the training data.
  In contrast, an interpretable LLM will often compromise performance by adding the constraint to only use a human-interpretable feature subset.
  Lastly, a regulatable LLM will further constrain this to be a feature set that is legally permissible.
  Naturally, these constraints will possibly lead to a degradation in performance.
  Note what there are exceptions, as e.g. what is considered interpretable can sometimes not degrade performance~\cite{kenny2023towards}.
  }
  \label{Fig:tradeoff}
\end{figure*}

In this paper, we are interested in the potential of interpretable ML to make models more regulatable.
Techniques from this field have been shown to help make models auditable~\cite{zhang2022explainable}, debug self-driving cars~\cite{kenny2024explainabledeeplearningimproves,dong2023did}, and calibrate appropriate trust~\cite{sanneman2022situation}.
However, to date there is no exploration of how to make interpretable LLMs for the purposes of regulation. 

In reality, regulation will likely take many different forms in different domains, but here we are specifically interested in the domain of insurance liability and how to regulate models in such a setting using interpretable ML.
In this domain, such institutions require their employees (and by extension their models) to use specific concepts in sensitive decisions in order to be legally compliant, but due to the black-box nature of AI, there is absolutely no way to verify this is happening~\cite{nguyen2021effectiveness}.
Hence, in these specific circumstances, a basic requirement for regulation is to force these models to use specific human-defined concepts in their inference process, which interpretable ML can help us do.
Interestingly, we find that in doing so, a dilemma presents itself in the form of a trade-off between regulation and performance previously unconsidered in the literature.
% However, it is worth noting that other domains may have different basic regulatory requirements, but here we are focused on insurance liability classification.

As an aside, we remind the reader that LLMs are broadly classified into two categories, generative (e.g., ChatGPT) and classification (e.g., BERT) models.
Although generative models have been at the forefront of recent attention, they are not the most practical for classification~\cite{zhong2023can,zhang2024pushing}.
In this paper, we focus on the classification type and use LLM to refer to them, which is common parlance~\cite{llm_2023}.

Next in Section~\ref{Section:related_work} we contextualize this work in the literature.
In Section~\ref{Section:tradeoff} we discuss context and the theoretical underpinning behind what we coin ``The Regulation Performance Trade-Off.'' 
In Section~\ref{Section:datasets} we describe the proprietary datasets used. 
Section~\ref{Section:method} describes our method for incorporating human-centered concepts into an interpretable LLM. 
Sections~\ref{Section:expts} and \ref{Section:UserStudy} describe experimental results, before our conclusion in Section~\ref{Section:conclusion}.

% For example, in insurance liability assessment one cannot rely on unfair or discriminatory features, there are specific concepts which \textit{must} be legally used in decisions to justify the final outcome.

% Moreover, the reasoning process also needs to be understandable, which motivates our usage of prototype-based reasoning~\footnote{Note this is also referred to as example-based reasoning or explanation-by-example~\cite{lipton2017precise}, exemplar-based reasoning~\cite{kim2014bayesian}, and case-based reasoning~\cite{leake1996case}. Here we use ``Prototype-based reasoning'' because we are using few examples.} to visualize the usage of these concepts, which has been shown to be understandable by end-users~\cite{kenny2021explaining}, useful to help decision making~\cite{chen2023understanding}, and preferred as as form of explanation over other popular approaches~\cite{jeyakumar2020can}.
% To the best of our knowledge, this idea to force the usage of human-defined concepts in CBR is a novel research direction, and moreover it is seen as one of the grand challenges of Explainable AI (XAI)~\cite{rudin2022interpretable}.

\section{Context and Trade-Off}
\label{Section:tradeoff}
As this paper focuses on the domain of insurance liability, this section gives some brief context in the area, before formalizing the regulation performance trade-off.

\subsection{Insurance Liability}
Explainable AI benefits from focusing on specific applications due to how it simplifies evaluation~\cite{XAI2024}.
Here we are focused on the specific task of determining liability in automotive accidents.
We want our system to (1) use human-vetted concepts in an interpretable way for regulation, and (2) benefit humans in a collaborative setting, both of which we show results for in our evaluation.
In insurance liability settings, there is an insured, and a claimant.
The insured is the person or entity that purchases an insurance policy from an insurance company, whilst the claimant is the person or entity that makes a claim for benefits under an insurance policy.
In our setting, the two parties are automotive drivers involved in a collision, and the accident is recorded in natural text, which motivates our usage of LLMs.
Legally required concepts to use in this domain consist of e.g. ``running a red light,'' and not other spurious (or even illegal) features such as e.g. a person's gender or nationality~\cite{benhamou2020artificial}.

\subsection{The Regulation Performance Trade-Off}
Without loss of generality, consider an LLM that encodes features into some latent space.
Within this, there exists a set of features which the LLM has learned to encode to perform optimally on some classification task, the \textit{"black box feature set"}.
There exists another set of features in the same space called the \textit{"interpretable feature set"}, which is the set of features humans can understand (e.g., a person's nationality).
In our case there also exists a final set, the \textit{"regulatable feature set"} (e.g., running a red light). 
This is a subset of the \textit{"interpretable feature set"}, as to be regulatable, a feature must be interpretable.
Formally, let $\mathcal{L}\in{\rm I\!R}^{(n)}$ be the $n$-dimensional latent space of the LLM. 
It follows that the sets are:
\begin{itemize}
    \item $\mathcal{B} \subseteq \mathcal{L}$: the \textit{"black box feature set"} that the LLM encodes to optimize a classification task.
    \item $\mathcal{I} \subseteq \mathcal{L}$: the \textit{"interpretable feature set"} that humans can understand.
    \item $\mathcal{R} \subseteq \mathcal{I}$: the \textit{"regulatable feature set"}, a subset of the interpretable feature set which allows legal usage of the LLM.
\end{itemize}
Thus, we have:
\[
\mathcal{R} \subseteq \mathcal{I} \subseteq \mathcal{L}
\]
\[
\mathcal{B} \subseteq \mathcal{L}
\]
The objective is to force the LLM to only use the set $\mathcal{R}$.
Note that $\mathcal{R}$ is not guaranteed to occupy the same space as $\mathcal{B}$, and is necessarily a subset of $\mathcal{I}$, given such constraints, a model relying only on $\mathcal{R}$ is guaranteed to have a performance equal to, or less than $\mathcal{B}$ or $\mathcal{I}$ (assuming $\mathcal{B}$ was trained well and we use $\mathcal{R}$ with the original LLM frozen).
Most important to note however, is that this illustrates how the interpretability performance trade-off [i.e., see~\cite{rudin2019stop}] is different.

\section{Insurance Datasets}
\label{Section:datasets}
The main datasets used in this paper originate from a global insurance company and are not publicly available.
However, in the spirit of scientific reproducibility, we also run our experiments on a publicly available and widely used dataset.
We briefly describe this latter dataset later in Section~\ref{Section:expts}, since it is already widely known and not our focus.

\subsection{The Liability Dataset}
This dataset contains 150,000 entries.
The columns are (1) natural language text statements describing a car accident between an insured and a claimant, and (2) a label from 0-100\% assigning liability to the insured, where 0\% is no liability and 100\% is complete liability.
To pre-process the dataset, we categorized liability into three classes: 

\begin{enumerate}
  \item \textit{Not Liable}: The insured is 0\% at fault in the accident.
  \item \textit{Split Liability}: The Insured and Claimant are both at fault (anywhere between 1-99\% at fault).
  \item \textit{Liable}: The insured is 100\% at fault
\end{enumerate}

After this, we balanced the dataset, which resulted in 14,000 entries for each class.
Furthermore, the data was divided into training (90\%), validation (5\%), and testing (5\%).

\subsection{The Human-Labeled Concept Dataset}
The second dataset is a collection of 2,000 statements, all of which are separate from the prior dataset.
For these data, we employed \textit{two separate vendors} to label parts of their sentences with important concepts for assessing liability that were defined by a domain expert.
Having two separate vendors is important because if our model were to have 45\% accuracy on classifying these concept labels, but the two vendors only agreed 60\% of the time, then it is actually a very good model having reached 75\% of this theoretical ceiling.
In total, there were eight labels (i.e., concepts) we asked them to assign, shown in Table~\ref{tab:concept_data}.
Both vendors precisely agreed on a given concept being present and its exact text within the statement 2.65\% of the time.
However, if we relax the second constraint and allow agreement when one text segment envelops the other, this agreement rises to 61.2\%, which we consider the ceiling of performance any model should realistically achieve.
For the final data, we joined all labels together from both vendors in order to maximize the amount of labeled concept data, so if Vendor 1 labeled the first ten statements with concept $x$, and Vendor 2 the last ten, we would collect 20 labels for that concept.

\begin{table}[!h]
    \centering
    \begin{tabular}{lc}
        \toprule
        Concept & Number of Labels \\
        \midrule
        IV Liable & 609 \\
        IV Not Liable & 501 \\
        IV Defensive Action & 503 \\
        IV No Defensive Action & 461 \\
        CV Liable & 712 \\
        CV Not Liable & 388 \\
        CV Defensive Action & 456 \\
        CV No Defensive Action & 501 \\
        \bottomrule
    \end{tabular}
    \caption{Human-Concept Dataset: The human centered concept dataset. There are eight concepts in total shown, with their corresponding number of labels in 2000 statements. IV = Insured Vehicle, CV = Claimant Vehicle.}
    \label{tab:concept_data}
\end{table}

The data can is summarized in Table~\ref{tab:concept_data}.
Notably, high-level concepts were chosen such as e.g. ``IV Liable'' rather than ``IV ran a red light'' to maximize the generalizability of the concepts during training.
We took 80\% of this data for training, and 10\% for validation and testing.

\section{Proposed Method}
\label{Section:method}
In this section we outline the assumptions for our proposed method of integrating human-centered concepts into LLMs, detail our architecture for doing so, and outline implementation specifics.

\subsection{Assumptions}
\begin{figure*}[!t]
  \centering
  \includegraphics[width=\textwidth]{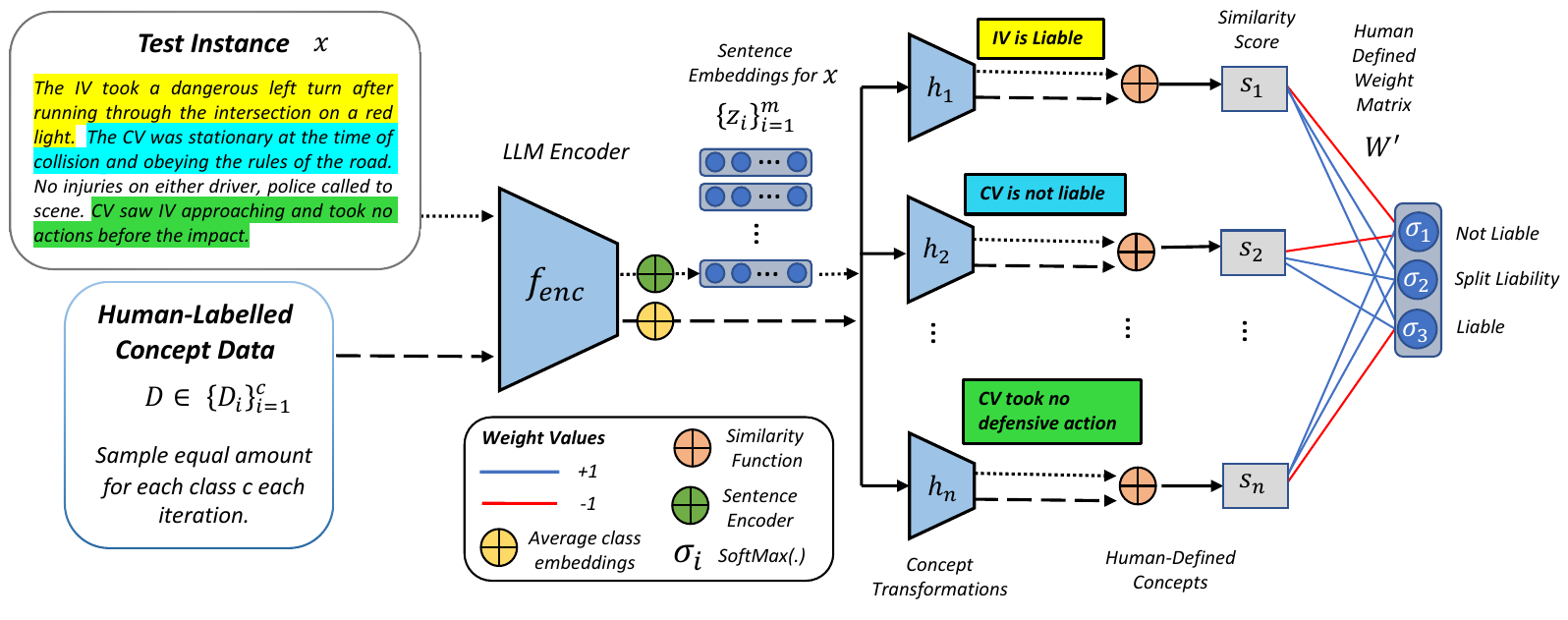} 
  \caption{
  Our proposed framework for regulatable LLMs: A test instance has its sentences encoded and compared to prototypes representing regulatable concepts defined \textit{a-priori} by humans.
  The maximum activation for each concept is used as similarity scores for the model's forward pass.
  Note, the test instance $x$ in this example is fabricated, it is \textit{not} an example of real data.
  % Note the figure is borrowed heavily from Kenny et al.~\shortcite{kenny2023towards}.
  }
  \label{fig:title}
\end{figure*}

We assume access to an encoder-only LLM trained for a specific classification task on a large quantity of data.
Furthermore, we assume access to (1) the original dataset used to train this LLM, and (2) another dataset of human-annotated concept data you wish to force the LLM to use during its classifications.
Lastly, we assume competent domain knowledge which can be used to define how each concept should contribute to each class.
For example, in our insurance liability domain, the concept ``IV Liable'' should positively contribute to the class ``Liable'', hence we manually define the classification weight matrix $W'$ to have a positive weight connection between this concept and class prediction, while it has a negative weight to the class ``Not Liable'' (see Figure~\ref{fig:title}).

\subsection{Architecture}
In the model shown in Figure~\ref{fig:title}, a test instance, $x$, is mapped to a set of sentence embeddings $Z\in\{z_i\}^m_{i=1}$ via the original encoder network $f_{enc}$ and a sentence encoder $\omega(\cdot)$.
Alongside this, a set of human-labeled sentence-level concept data $D$, which can be divided up into each concept class $D\in\{D_i\}^c_{i=1}$ is also passed into $f_{enc}$ to produce a set of embeddings $D_c\in\{d_i\}^k_{i=1}$ for each class.
These $c$ sets are then averaged into $c$ concept prototypes $P\in\{p_i\}^c_{i=1}$, one for each concept $c$.
Then, for example, all of the sentence embeddings for $x$ (i.e., $Z\in\{z_i\}^m_{i=1}$) and prototype $p_i$ are passed into $h_i$ to measure each sentence's similarity to $p_i$ via a similarity function, before its element-wise product with $W'$ is taken to produce the network's logits with:
\begin{equation}
\label{eq:sim}
\texttt{sim}(z_{i}, p_{i}) = \log\left(\frac{(z_{i} - p_{i})^2 + 1}{(z_{i} - p_{i})^2 + \epsilon}\right)
\end{equation}
\begin{equation}
\label{eq:argmax}
s_i = \arg\max_{z_i \in Z} \quad \texttt{sim}(z_i, p_i)
\end{equation}
\begin{equation}
\label{eq:logits}
\hat{y} = \vec{s} \odot W'
\end{equation}
\noindent where $\vec{s}$ is the vector of similarity scores  for each concept such that $\vec{s}=\{s_1, s_2,...s_n\}$, and $\epsilon$ is to avoid division by zero.
Equation~\ref{eq:sim} is  monotonically decreasing such that if the prototype is close to a sentence embedding, it will output a high similarity score.
The maximum similarity score across all sentences is then used in the forward pass for that concept, and this is repeated to give a score for all concepts with Equation~\ref{eq:argmax}.
Finally, this vector of similarity scores takes an element-wise product with $W'$ in Equation~\ref{eq:logits} to give the logits $\hat{y}$.

The loss of our network is calculated with two terms, the first $\mathcal{L}_c$ is a standard loss for the class label, and the second loss $\mathcal{L}_h$ is the human-concept loss.
For $\mathcal{L}_h$, a subset of each concept label $D'\in\{D'_i\}^c_{i=1}$ is passed each iteration into their corresponding $h$, and their similarity scores against the pre-computed prototypes that same iteration are calculated with Equation~\ref{eq:argmax} for a cross-entropy loss.
Together, this has the effect of encouraging the network to classify the overall label correctly, but also to learn to classify the human-concept data correctly with the prototypes, which together enforces the necessary constraints for our system.
The loss can be written as:
\begin{equation}
    \label{eq:process}
    \min_{ \phi, \omega, W' } \mathcal{L}_c( y, \hat{y} ) +
     \frac{1}{C}\sum_{i=1}^{C} \mathcal{L}_h( y', \phi(p_i, D'_i)  ) 
\end{equation}
\noindent where $y$ is the overall label, and $\hat{y}$ is the prediction of the overall label.
Moreover, $\phi(\cdot)$ is a function that outputs a vector of similarity scores $\vec{s}$, $y'$ is the label for the human concept, $p_i$ is the computed prototype for concept $i$ that iteration, and $D'_i$ is randomly sampled concept data for concept $i$.
Put simply, the first term teaches our network to predict the right class, and the second encourages it to learn to classify concepts correctly with the prototypes.

% \paragraph{CBR Explanation.} As an aside, the closest training human concept labelled datapoint to each sentence can also be used as a CBR explanation by saying e.g. \textit{``I think this sentence has the concept `IV Liable' because it is very similar to this labelled example [...show example to user...] stored for this concept''.}
% This allows us to combine both concept-based explanation and CBR explanation, something not considered in the literature before but seen as one of the grand challenges of XAI~\cite{rudin2022interpretable}.

\subsection{Implementation Details}
\label{Section:implementation_details}
To encode a set of sentence embeddings with $\omega(\cdot)$ there are two main ways we explore, \textit{context unaware} and \textit{context aware}.
For \textit{context unaware}, we break the input text $x$ into sentences prior to encoding with $f_{enc}$, and use the BERT [CLS] token (or equivalent) as the sentence embeddings.
For \textit{context aware}, we pass all of $x$ into $f_{enc}$, divide up the contextualized word embeddings (i.e., the token embeddings after the forward pass) into sentences, and then collate them into a single embedding.
In our experiments, to collate these we used either (1) a simple average, (2) a recurrent neural network (RNN) encoder, or (3) an attention layer.
% This function has the effect of taking the contextualized word embeddings for each sentence (which are all different lengths representing the words composing the sentence), and representing them all as a single vector each, which is needed for our model.

The transformations $h_i$ are all MLP networks with one hidden layer.
To regularize, we compressed the dimensionality here to as low as possible without compromising performance.
For our experiments, this involved going from an encoding space of size 768 to 16 in these MLP networks.

Lastly, for $W'$, expert knowledge is needed to define it appropriately.
In our case, we used domain knowledge from an industry expert and assigned either +1 or -1 to the weight connections prior to training.
We allowed the model to fine-tune these weights during training, but only the magnitude was allowed to change, not the sign/polarity (e.g., a +1 weight will change to 0.9 or 1.1 during training, but not -0.5).
This ensured (for example) that the concept ``IV Liable'' would always positively contribute to the class ``Liable''.

At testing time, the entirety of the human-concept dataset for each concept is averaged into a single prototype for each concept and cached.

\section{Computational Experiments}
\label{Section:expts}
Here, we describe our baselines, before detailing the datasets, metrics, and finally the results.

\begin{table*}[t!]
\resizebox{\textwidth}{!}{%
\begin{tabular}{@{}lllrrrrr@{}}
\toprule
\multicolumn{8}{c}{Context Unaware}                                                                                                                                                                                                       \\ \midrule
             & \multicolumn{1}{l|}{}                  & \multicolumn{3}{c|}{Insurance Liability Data}                                                  & \multicolumn{3}{c}{Beer Advocate Dataset}                                        \\ \midrule
Human Labels & \multicolumn{1}{l|}{Sentence Encoding} & \multicolumn{1}{c}{Acc.} & \multicolumn{1}{c}{Top 1} & \multicolumn{1}{c|}{Top 3}              & \multicolumn{1}{c}{Acc.} & \multicolumn{1}{c}{Top 1} & \multicolumn{1}{c}{Top 3} \\ \midrule
Yes          & \multicolumn{1}{c|}{-}                 & 60.75±0.14               & \textbf{45.90±0.11}       & \multicolumn{1}{r|}{\textbf{75.9±0.27}} & 77.41±0.24               & \textbf{44.32±0.23}       & \textbf{74.43±0.16}       \\
No           & \multicolumn{1}{c|}{-}                 & 63.29±0.05               & 7.27±0.24                 & \multicolumn{1}{r|}{28.63±0.23}         & 80.07±0.05               & 8.75±0.17                 & 26.32±0.10                \\ \midrule
\multicolumn{8}{c}{Context Aware}                                                                                                                                                                                                         \\ \midrule
             & \multicolumn{1}{l|}{}                  & \multicolumn{3}{c|}{Insurance Liability Data}                                                  & \multicolumn{3}{c}{Beer Advocate Dataset}                                        \\ \midrule
Human Labels & \multicolumn{1}{l|}{Sentence Encoding} & \multicolumn{1}{c}{Acc.} & \multicolumn{1}{c}{Top 1} & \multicolumn{1}{c|}{Top 3}              & \multicolumn{1}{c}{Acc.} & \multicolumn{1}{c}{Top 1} & \multicolumn{1}{c}{Top 3} \\ \midrule
Yes          & \multicolumn{1}{l|}{Mean}              & 66.28±0.94               & 19.77±0.12                & \multicolumn{1}{r|}{50.9±0.76}          & 81.40±0.63               & 18.81±0.81                & 54.08±0.35                \\
Yes          & \multicolumn{1}{l|}{RNN}               & 63.87±0.27               & 14.09±0.67                & \multicolumn{1}{r|}{35.9±0.44}          & 83.06±0.99               & 13.04±0.23                & 33.62±0.91                \\
Yes          & \multicolumn{1}{l|}{Attention}         & 64.52±1.12               & 17.27±0.55                & \multicolumn{1}{r|}{46.13±0.32}         & \textbf{85.05±0.12}      & 20.42±0.78                & 51.13±0.71                \\
No           & \multicolumn{1}{l|}{Mean}              & \textbf{69.01±0.83}      & 12.27±0.41                & \multicolumn{1}{r|}{33.86±0.22}         & 83.72±0.37               & 15.11±0.99                & 35.18±0.57                \\
No           & \multicolumn{1}{l|}{RNN}               & 68.10±0.68               & 10.22±0.29                & \multicolumn{1}{r|}{40.45±0.89}         & 80.40±0.18               & 6.61±0.45                 & 24.81±0.04                \\
No           & \multicolumn{1}{l|}{Attention}         & 67.84±0.35               & 15.01±0.81                & \multicolumn{1}{r|}{37.95±0.76}         & 83.72±0.51               & 13.51±0.63                & 33.33±0.92                \\ \bottomrule
\end{tabular}}
\caption{Computational Results: 
The best results were achieved by supervising with human-labelled concept data (i.e., Human Labels=Yes), and using context unaware sentence embeddings.
This resulted in lower accuracy on the class label compared to unsupervised baselines (i.e., Human Labels=No) as predicted in Section~\ref{Section:tradeoff}.
Best results are in bold. 
Note the original black-box accuracy was 68.68\% and 84.16\% for the Insurance Liability and Beer Advocate datasets, respectively.
Standard error across three iterations is shown alongside the results.
}
\label{tab:example}
\end{table*}

\subsection{Baselines}
% To the best of our knowledge, there is no work which allows the integration of human-centred concepts and prototypes as we have done, particularly on the sentence level. 
% Hence, w
We conduct comparisons between our regulatable model in Figure~\ref{fig:title} and a generic baseline which does \textit{not} use human-centered data (i.e., Human Labels=No in Table~\ref{tab:example}).
These unsupervised baselines set the prototypes as learnable parameters instead, which is representative of the literature~\cite{chen2019looks,antognini2021rationalization,ming2019interpretable,das2022prototex}.
Alongside this we also randomize $W'$ and don't constrain its polarity in baselines to avoid any human bias making its way into the training process.
While comparing these two baselines, we also do so in (1) a context aware fashion, and (2) a context unaware one (see Section~\ref{Section:implementation_details}).
For our text encoder we use BERT~\cite{devlin2018bert}, note we tried a grid search of several other architectures such as DeBERTa, RoBERTa, DistilBERT etc., but none showed a significant improvement, so we used BERT because it is the most widely researched.

\subsection{Datasets}
Our primary tests are on the insurance liability datasets detailed already in Section~\ref{Section:tradeoff}, as we are particularly interested in evaluating our technique on real-world applications.
However, to foster reproducibility, we also extend the same tests to the Beer Advocate dataset~\cite{mcauley2012learning}.
This second dataset is 200k rows of text data detailing reviews of beers, it contains the concepts of \textit{Appearance}, \textit{Aroma}, \textit{Palate}, \textit{Taste}, and \textit{Overall}.
To mimic related work~\cite{bao2018deriving}, we divide the dataset into a binary classification problem of those reviews with a score higher than 4, and lower.
The Beer Advocate dataset is also quite unique in that it contains 994 sentence-level annotations for the five concepts present, making it suitable for our needs.
We further divided these concepts into positive/negative ones (depending on which class they belonged to) to make in total 10 concepts which could be used for classifying the positive/negative reviews.
Going forward, we will talk about \textit{class labels} (i.e., the regular classification task), and \textit{concept labels} (i.e., the sentence-level annotations), as they are two different evaluations.

\subsection{Metrics}
We consider three primary measurements.
First, we measure how well the models are performing on their respective \textit{class labels}.
Following best practice, a model is chosen based on its performance on validation data during training, and then performance on the testing data is reported.
Next, we also consider how well the model is classifying the \textit{concept labels}.
For this we consider a ``Top 1'' and ``Top 3'' metric, the model is seen as correct if the prototype for e.g. ``IV Liable'' activates the strongest for a \textit{sentence} in a datum with that label (i.e., Top 1 metric), and likewise for Top 3 it is seen as correct if it is in the 3 most strongly activated.

\subsection{Results}
Table~\ref{tab:example} shows the results of running our tests three times and calculating the mean alongside standard error.
Overall, there are three strong trends to note.
Firstly, the context aware setting achieves better classification performance on the class labels, whilst the context unaware models do better at classifying the concept labels.
This is likely because the latter forces the LLM to have stronger sentence representations that are not entangled with the rest of the text, this works better for concept classification.
Secondly, there is another strong trend that learning the concept representations from scratch instead of using the labels (i.e., Human Label=no) results again in stronger classification performance of the class, but again this comes alongside a trade-off with concept accuracy.
Thirdly, the attention mechanism in context aware settings does best at encoding sentence representations when compared to taking an average or using an RNN.

The strongest results come from the context unaware model using the human annotated data.
This model achieved 45.90±0.11 / 75.9±0.27 Top 1/Top 3 classification performance on the concept labels for the Insurance Liability dataset, respectively, and 44.32±0.23 / 74.43±0.16 Top 1/Top 3 classification performance on Beer Advocate, respectively.
Importantly however, this did come with a trade-off on performance for the actual overall class labels.
Specifically, on the Insurance Liability data the performance dropped from the initial black-box model accuracy of 68.68\% to 60.75\%, and on Beer Advocate from 84.16\% to 77.41\%, resulting in an average drop of 7.34\% in performance.
In contrast, the models which are not confined to regulatable features and instead learned the interpretable concepts actually outperformed the original black-box, reaching 69.01\% on the Insurance Liability data, and 85.05\% on Beer Advocate.
This improved performance could be attributed to a regularization effect induced by our model, which forces the LLM to reason using only a handful of prototypes, as similar results were seen before with similar techniques~\cite{kenny2023towards}.
Recall that the inter-rater reliability, as measured by the percentage agreement between human raters, was 61.2\% for the insurance data concept labeling task (see Section~\ref{Section:datasets}). Consequently, the reported results actually reach 75\% of this theoretical ceiling.
Most importantly however, this lends a noteworthy datapoint which helps to quantify the trade-off between regulatory constraints and performance in LLMs whenever transparency of concept usage in classifications is required.

% As a side note, while a concept classification accuracy of 43.40\% and 44.32\% may seem insufficient, there are two crucial things to note.
% Firstly, these results are in line with prior work showing similar (or indeed worse) classification performance on Beer Advocate~\cite{conrat}.
% Secondly, recall the Insurance Liability concept data was labelled by two separate vendors, which commonly disagreed.
% This is important because, for example, the two vendors only agreed on a concept being present, and the exact text for this concept 2.65\% of the time.
% However, if we relax the second constraint and allow agreement when the text segments overlap, this agreement raises to 61.2\%.
% Overall, this illustrates how concepts are somewhat subjective and overlap commonly, which motivated our usage of the Top 3 metric.
% This problem is exasperated on Beer Advocate which was only labelled by a single user.

% \subsection{Data Ablation}
% Go into results of how results are also due to low training data most likely.
% We are looking into how the amount of labelled data affects the outcome.

\section{Pilot User Study}
\label{Section:UserStudy}
Here we facilitate an ``Application Grounded Evaluation,'' which is typically seen as the gold-standard in explainable AI~\cite{doshi2017towards}.
Specifically, we recruited eight adjusters from a private global insurance company (whose full-time job it is to process insurance claims) to participate in a pilot study using our model to help classify real statements in practice.
While this meant our sample size would be necessarily reduced, it allowed the enormous advantage of using real-world data in a real-world setting.
Studies have consistently shown that how users react to AI technology is quite divided~\cite{brecheisen2024}.
Given this, our hypothesis was that certain users would react favorably to the AI and cluster into one group with reduced time taken overall to classify the statements, whilst the others would do the opposite.

\paragraph{Materials.} 
We designed a within-subjects study which showed adjusters eight separate statements, four with AI assistance and four without.
The questions with AI assistance showed adjusters one concept activation per statement, which was most relevant.
Adjusters were told all statements could be either liable, split liability, or not liable.
However, in reality, four were liable, and four not liable, with the AI assistant helping on half of each.
The eight adjusters were split into two groups, in which the questions with AI assistance were counterbalanced.
The AI assisted questions gave (1) its prediction for the statement, and (2) the highlighted text for the most important sentence in the prediction.
The final analysis pooled all data from both versions of the survey together to control for the effect of each individual question.
Each participant was given the survey online and asked to complete it in their own time (but during working hours), in one sitting.
The study passed IRB review.

\paragraph{Metrics.} 
We measured (1) how accurately each adjuster classified each statement, (2) how quickly they classified each statement, and (3) how confidently they classified each statement.
The confidence metric was measured on a 7-point Likert scale with the question \textit{``I am confident in this classification''}.
Each user's scores for statements with and without the AI assistant were averaged into a single result, giving two measurements for each metric for each user as standard~\cite{kenny2021explaining}.

\subsection{Results} 
\begin{figure}[t!]
  \includegraphics[width=\columnwidth]{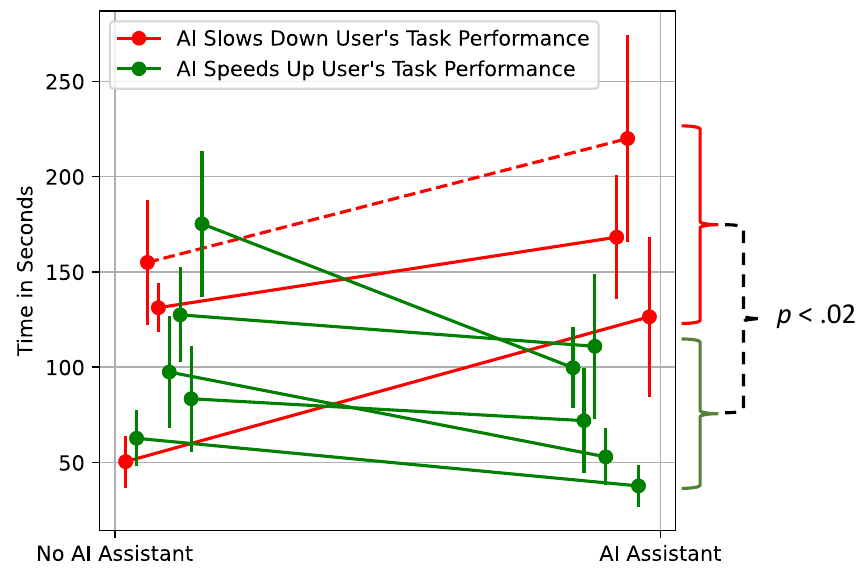}
  \caption{Time Results: Each user's average time to complete statements with and without the AI assistant is shown. 
  Statistically significant results were seen in those users who benefited form the AI against those who did not, with both forming two distinct clusters regardless of their baseline without the AI.
  Standard error shown.
  The dashed line represents User 3 who seemed averse to the AI overall.}
  \label{Fig:user_study}
\end{figure}

First, the data was cleaned (details in Appendix~\ref{sec:appendix}).
Overall, our hypothesis was confirmed when we found user scores on \textit{time taken} became widely divergent based on how they responded to the AI (note Figure~\ref{Fig:user_study}).
Those users whose time got longer with the AI ($n$=3) vs. those users whose time got less ($n$=5) saw a statistically significant difference (tested for normality; $t$(6)= 3.59, $p$ $<$ 0.02).
Overall, even if we pool both groups together, this still averaged as 110.40 $\pm$ 14.61 seconds with the AI assistant compared to 123.46 $\pm$ 29.61 without, hinting towards a benefit of the AI assistant on a population level.
On confidence scores, a similar trend was seen in users whose confidence improved with the AI ($n$=3) and those whose got worse (n=$3$; $t$(4)=3.59, $p$=0.094).
% These results are echoed when examining people's confidence, where 5/8 users showed no difference or improved confidence when using the AI assistant, whilst 3/8 showed lower confidence with the AI assistant.
Overall, this averaged at 6.5 $\pm$ 0.27 with the AI assistant compared to 6.4 $\pm$ 0.42 without it.
Given the average confidence was so high overall, this represents a notable increase.

As an interesting aside, only User 3 made a mistake when classifying the statements (see dashed line in Figure~\ref{Fig:user_study}). 
Specifically, they classified the second question as ``Split Liability'' when it was ``Liable''.
This question for the user had an AI assistant, indicating a possible lack of trust towards the AI, as all other participants agreed with the AI on this question.
Note this user spent the longest time deciding on classifications with the AI, lending evidence that a lack of trust in AI contributes to slower task performance.

In summary, this study indicates two intriguing findings.
Firstly, despite the regulatory model having worse performance compared to a black-box on class labels, humans still benefit overall from interacting with it, as indicated by their improved speed.
Moreover, because adjusters were almost always correct in their classification, their improved confidence score with the AI was also \textit{appropriate confidence}, similar to the idea of \textit{appropriate trust} in AI~\cite{sanneman2022situation}.
Secondly, as prior work has hinted~\cite{brecheisen2024}, how people respond to the AI assistant is quite individual, but if those users who benefit can be identified pre-hoc, the system's potential utility increases.

% an intriguing finding that how people react and use the technology will be quite individualistic, but those who benefit overall will cluster into a significantly better group than those who do not, \textit{regardless of their ability without the AI}.
% Importantly, similar to the idea of \textit{appropriate trust} in AI~\cite{sanneman2022situation}, because users were quite accurate in their classifications, this is \textit{appropriate} confidence, rather than inappropriate.

% \begin{figure}[t!]
%   \includegraphics[width=0.48\textwidth]{imgs/conf.pdf}
%   \caption{}
%   \label{Fig:user_conf}
% \end{figure}

\section{Related Work}
\label{Section:related_work}
Regulation in machine learning has come into the spotlight recently, with major conferences dedicating workshops to the topic~\cite{ma_0}, governments trying to implement it~\cite{wischmeyer2020regulating}, and academia actively researching it~\cite{onitiu2023ai}, but there is little work on how it should be concretely realized.
Due to this sparsity, in our literature review, we focus on tangential work which has built inherently interpretable LLMs, as it is widely agreed to be a prerequisite for AI regulation~\cite{casper2024black}.

Case-based reasoning (CBR) for interpretable LLMs is a recent idea, it uses real examples from the training data directly in inference for interpretability purposes.
Notable work in this area can be traced back to Ming et al.~\cite{ming2019interpretable} who focused on RNNs.
Das et al.~\cite{das2022prototex} proposed ProtoTEx, which classifies test instances with reference to learned prototypes (i.e., examples or ``cases'').
Van Aken et al.~\cite{van2022patient} proposed ProtoPPatient, which works for multi-label classification.
Xie et al.~\cite{xie2023proto} is the most up to date work, which adds saliency maps to the explanation. 
Similar work exists in the concept explanation literature~\cite{chan2022frame,bouchacourt2019educe,antognini2021rationalization}.
In contrast to all these, our work allows the direct integration of human-regulatable concepts into the inference process, which is needed for the type of regulation we are striving for.
% As an aside, all this work also bears resemblance to concept-bottleneck models~\cite{koh2020concept}, but in contrast our approach allows the visualization of the concepts (and the usage of prototypes), which is better for transparency.

Perhaps the most closely related work is that of Kenny et al.~\shortcite{kenny2023towards}.
The authors proposed to explain a deep reinforcement learning agent by wrapping its encoder with an interpretable prototype layer, where each prototype represents a human-friendly concept, but the authors note the networks are prone to over-fitting, likely because they only use a single example to represent each concept.
We build upon this work by collecting a large human-annotated dataset for each concept to avoid over-fitting, and adapting the framework for LLMs.
% This latter point is important as Kenny et al.~\shortcite{kenny2023towards} worked with ``whole'' cases for explanations, whilst in our case we are breaking each down into individual sentence embeddings (see Figure~\ref{fig:title}), which is a more intricate learning process.

% Lastly, we contextualize our work within the mechanistic interpretability literature~\cite{nanda2023progress}.
% In this area, one of the core challenges is superposition, where single neurons in LLMs represent many features simultaneously~\cite{bereska2024mechanistic}.
% Recent work by Zimmermann et al.~\shortcite{zimmermann2024scale} showed that as LLMs get bigger, this problem gets worse, and the authors concluded the need for \textit{monosemanticity} (i.e., making single neurons represent single features/concepts) to be integrated into the LLM with intent \textit{pre-hoc}.
% Recent posts by Anthropic and OpenAI have reported achieving some separation in an unsupervised manner by training sparse auto-encoders to isolate features of interest which can manipulate the LLM outputs~\cite{bricken2023towards,templeton2024scaling}.
% In contrast to all this, we disentangle features using human labels to allow single neurons to represent dedicated human-defined concepts.

\section{Discussion \& Conclusion}
\label{Section:conclusion}
In this paper, we proposed a framework for helping to regulate LLMs.
Our primary goal was to instantiate a regulatable LLM in insurance liability settings and quantify the trade-off (if any) which occurs related to performance and user interaction.
Results showed that one can constrain an LLM to use regulatable concepts post training, but that this does degrade performance by around 7.34\% on average, an interaction we coin as ``The Regulation Performance Trade-Off.''
However, given that it is currently impossible to deploy these models in many sensitive applications due to their black-box nature~\cite{rudin2019stop}, this will often be a small price to pay.
More importantly though, our user study with industry professionals highlighted the positive utility of the method in practice for human-AI collaboration despite this trade-off, which is a sobering reminder that the model's performance on class labels is only part of the overall picture to be considered in evaluation.
We hope these data will take the world a step closer to regulatable LLMs that benefit humanity.

\section*{Limitations}
Here we detail the limitations of our work which give way to opportunities for future research.

\paragraph{LLM Constraints.}
Our model is limited to the learned representations of the original LLM.
It could be that by training end-to-end, the results would be superior, but our preliminary experiments failed to accomplish this.
It would however be interesting to explore this in future work as a way to achieve superior representations for the human-centered concepts.

\paragraph{Small Sample.} 
Our user study design opted for a smaller sample size in order to test it with real industry professionals in a realistic deployment setting. 
This has the huge advantage of truly testing the system ``in the wild'', but comes with the trade-off of a small sample of users.
Hence, even though our test reached statistical significance, it should be verified on a larger sample of end users.

% \paragraph{Generalizability.}
% Currently it is not entirely clear how to apply this to e.g. images or generative AI.
% Our method breaks down a text input into sentence-level embeddings, so any attempt to generalize the method would require consideration of how to do this in images or GenAI.

% \paragraph{Regulation Definition.}
% It is currently not agreed what will definitely constitute effective regulation of AI.
% However, most entities and individuals agree that some level of transparency will be required.
% A limitation of our method is that we have assumed this will contribute to regulation, but until this is more strictly defined by governments and other legal entities, we do not know for sure.

\paragraph{Separation of Explanation and Prediction.}
It is not clear from our user study design if the explanation or model prediction made the core difference in the study.
As the AI assisted questions showed both the AI prediction and the concept explanation, it is not clear which made a difference.
This is a common issue however~\cite{lundberg2018explainable,barnett2024improving}, as such studies are so expensive to run, and naturally have so few users, it is often an unfortunate necessity to avoid splitting the user base into so many conditions that the results become impossible to interpret.

\paragraph{Labeling Requirements.}
Our method requires a large dataset of human-annotated concepts.
This is a large bottleneck for the method, but it is conceivable that generative language models could actually be made to synthesize this data, which would be interesting to investigate in future research.

\paragraph{Generalizing.}
Our method is developed for encoder-only language models. 
It may require several alterations to make similar methods work for decoder-only language models or image classifiers.

\bibliography{acl_latex}

\appendix

\section{Appendix}
\label{sec:appendix}

\subsection{User Study Data Cleaning}
First, we found two outlier entries which were excluded from analysis.
Specifically, one user spent over 10x times longer to complete one question compared to all other entries in the dataset (including their own other questions), so this was excluded assuming the user was momentarily distracted.
Additionally, one user logged a confidence score of 1 for their final question, when the lowest score in the data overall otherwise was 4, the average $>$ 6, and indeed the user in question logged 6 as their lowest score otherwise.
Note we only excluded the specific metric on the specific question for the specific user, all the user's data otherwise was included as normal.

\subsection{Computational Budget}
We train our models on 4 GPUs using AWS, to reproduce the results would take 1 day on average.

\subsection{User Study Design}
Here we post the entire user study, as much as possible, for transparency.

\begin{figure}[t!]
  \includegraphics[width=\columnwidth]{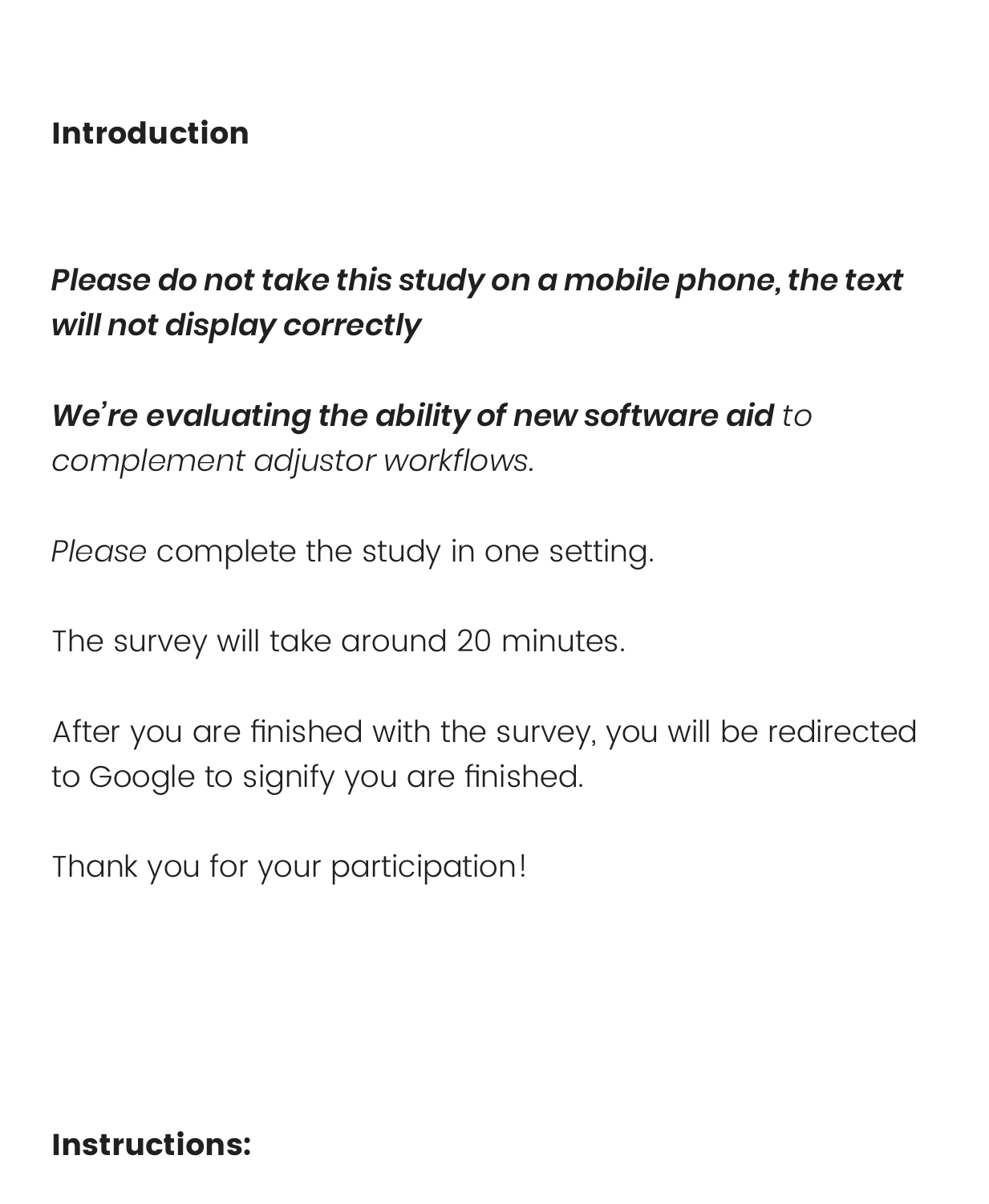}
  \caption{Page 1 of user study}
\end{figure}

\begin{figure}[t!]
  \includegraphics[width=\columnwidth]{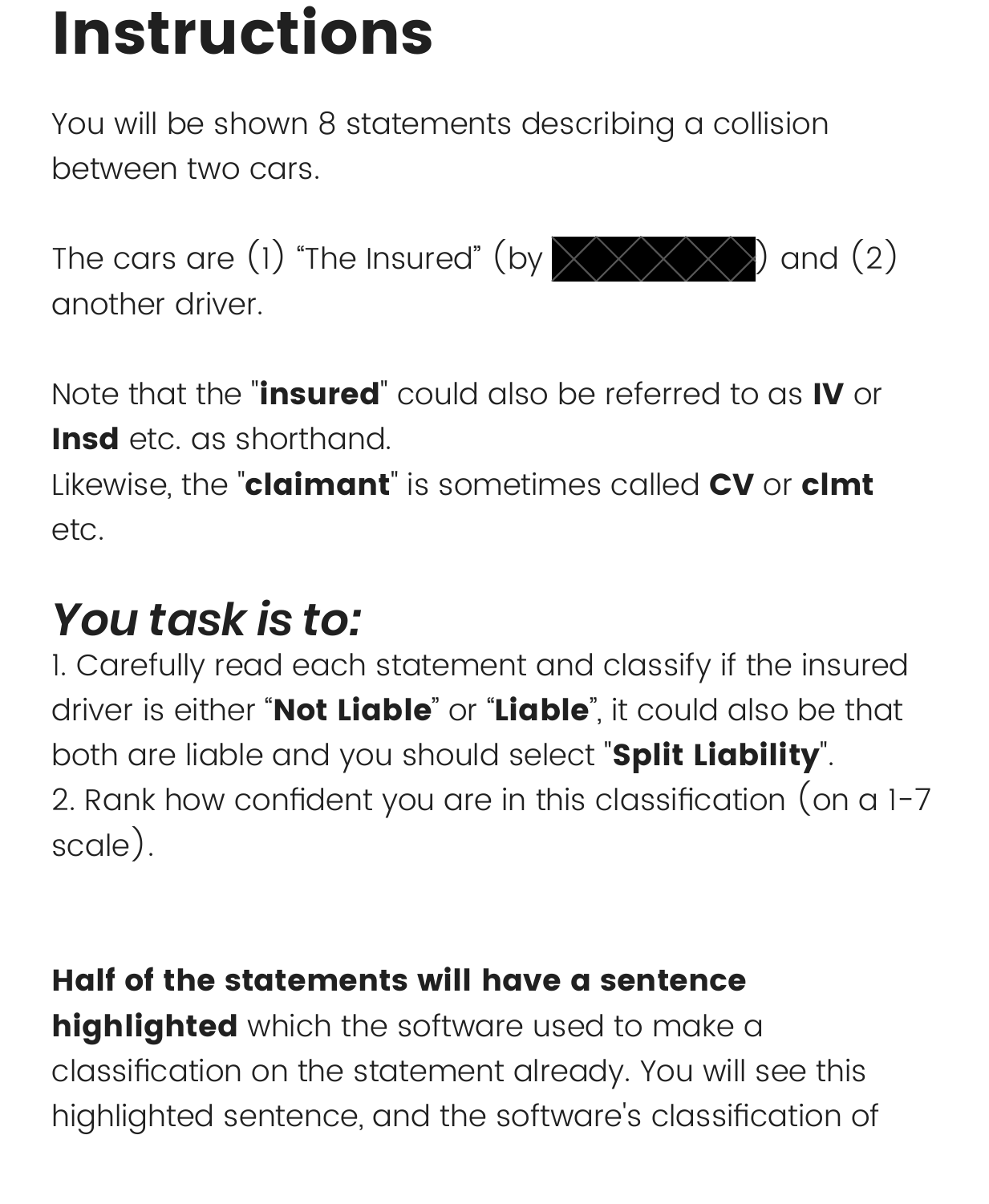}
  \caption{Page 2 of user study}
\end{figure}

\begin{figure}[t!]
  \includegraphics[width=\columnwidth]{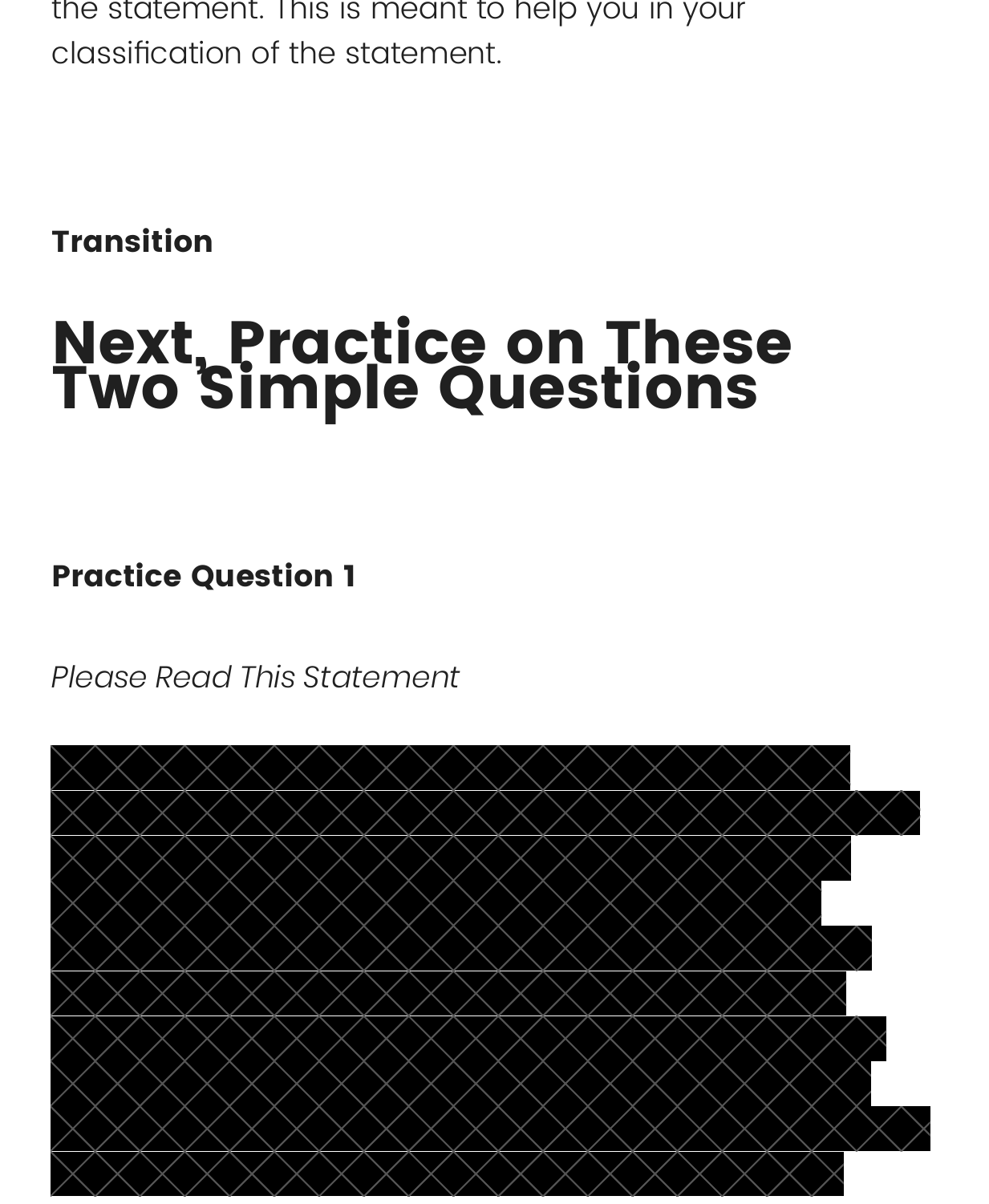}
  \caption{Page 3 of user study}
\end{figure}

\begin{figure}[t!]
  \includegraphics[width=\columnwidth]{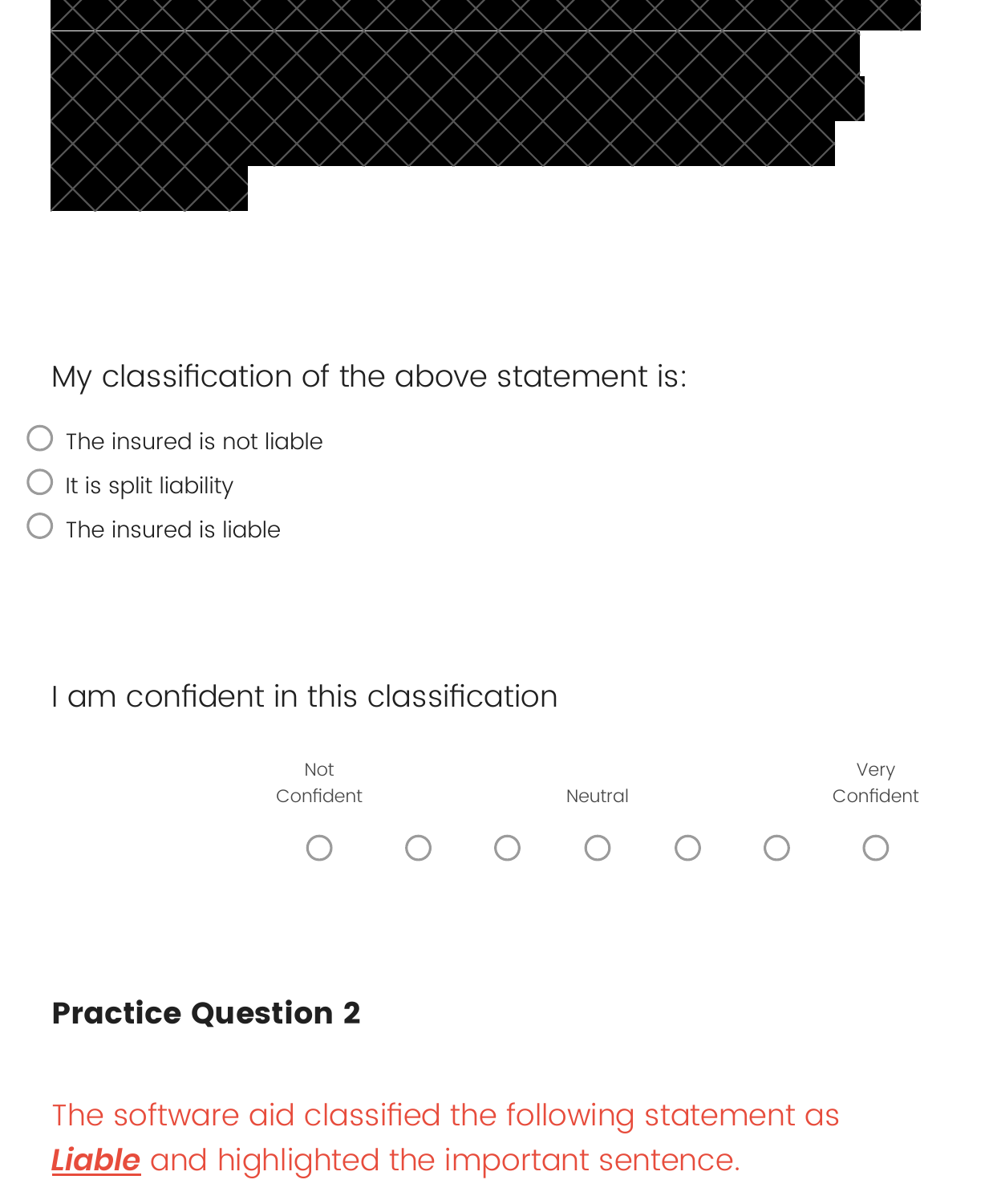}
  \caption{Page 4 of user study}
\end{figure}

\begin{figure}[t!]
  \includegraphics[width=\columnwidth]{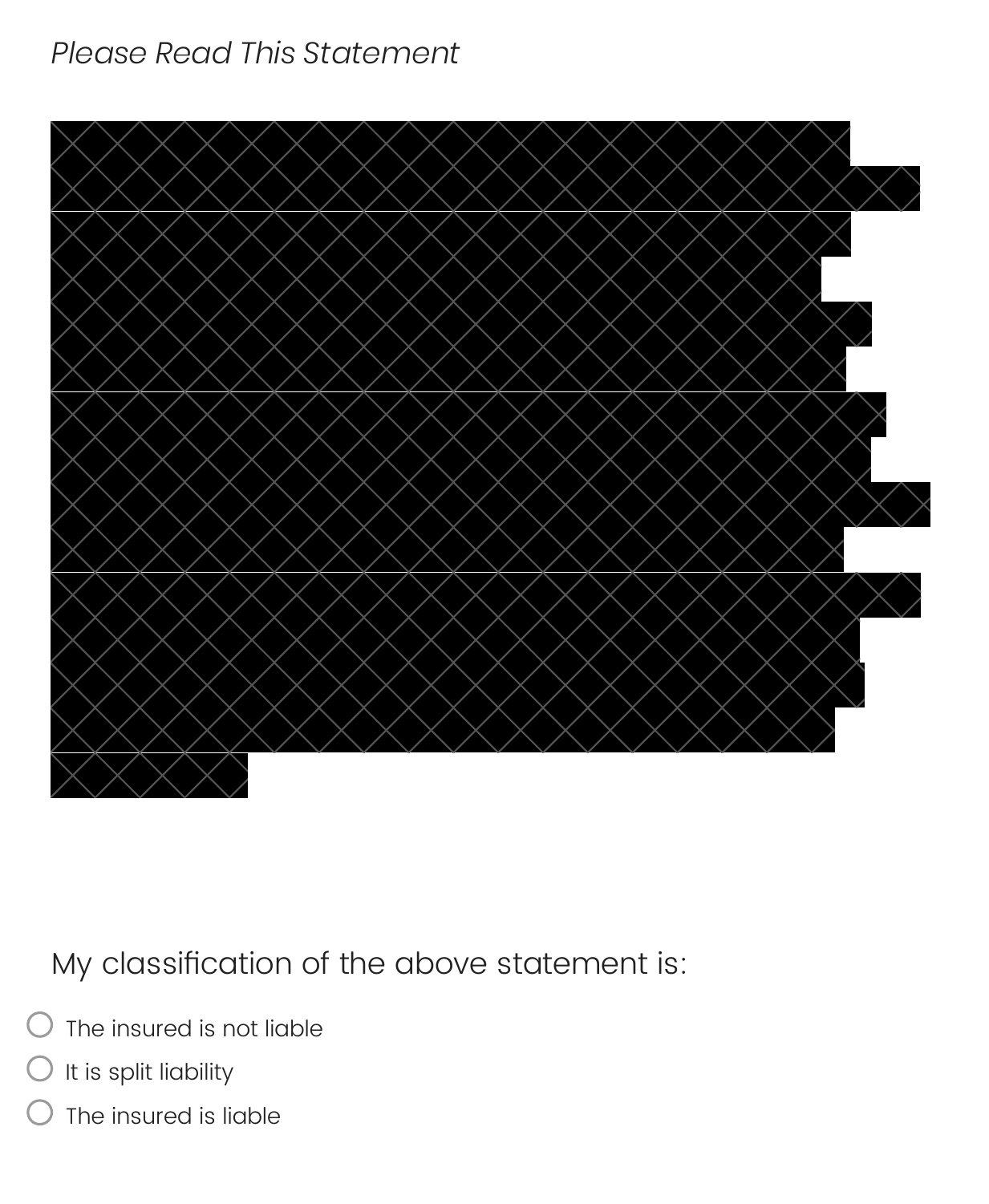}
  \caption{Page 5 of user study}
\end{figure}

\begin{figure}[t!]
  \includegraphics[width=\columnwidth]{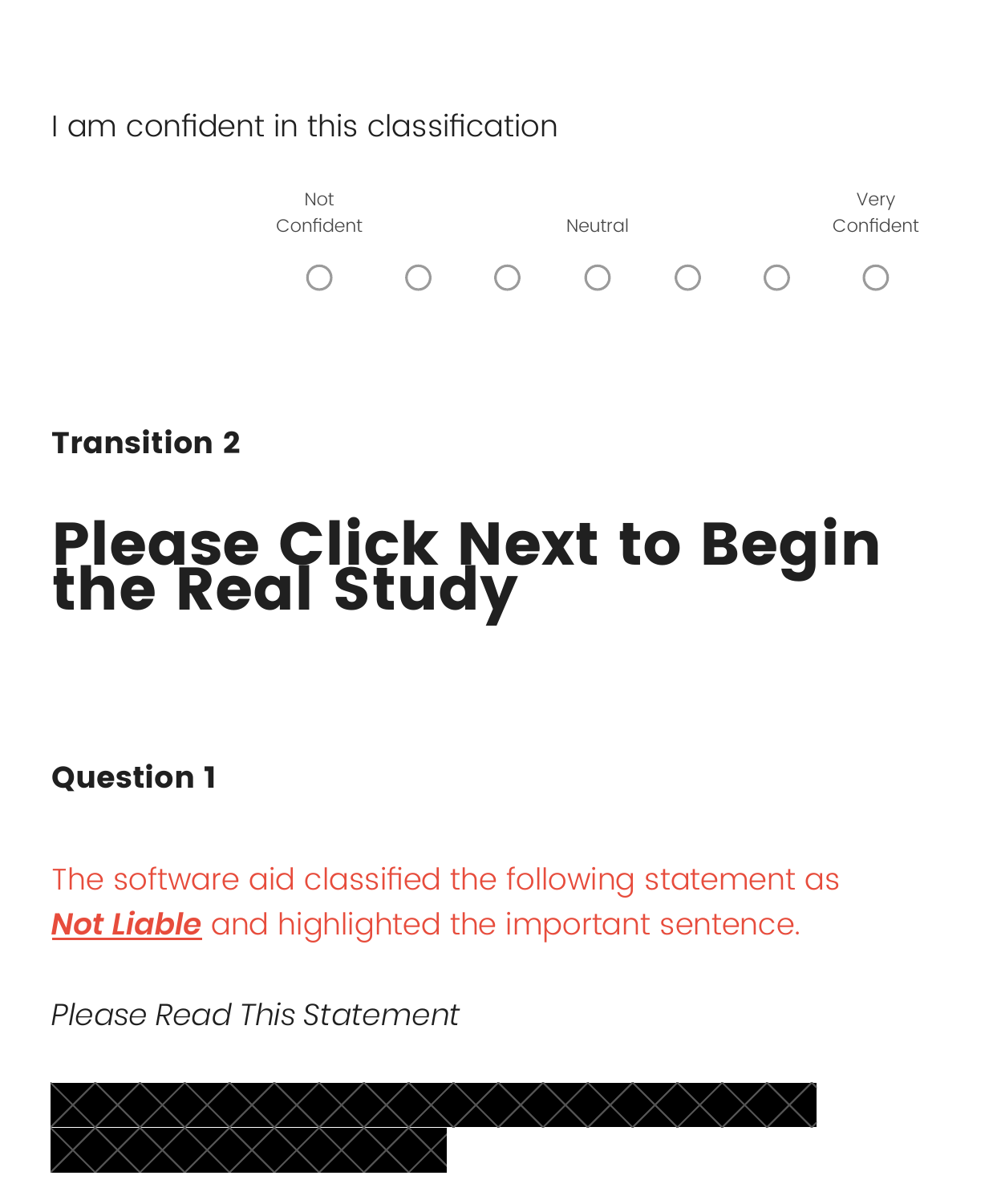}
  \caption{Page 6 of user study... (study is repetitive after this and omitted).}
\end{figure}

\end{document}